\documentclass[11pt]{article}

\usepackage[final]{acl}

\usepackage{times}
\usepackage{latexsym}
\usepackage{amsmath}
\usepackage{amssymb}
\usepackage{amsthm}
\usepackage{algorithm}
\usepackage{algorithmic}
\usepackage{booktabs}
\usepackage{multirow}
\usepackage{mdframed}   
\usepackage{hyperref}

\usepackage[T1]{fontenc}

\usepackage[utf8]{inputenc}

\usepackage{microtype}

\usepackage{inconsolata}

\usepackage{graphicx}

\usepackage{xspace}
\newcommand{\ours}{\textsc{SESS}\xspace}

\title{Submodular Evaluation Subset Selection in Automatic Prompt Optimization}

\author{
Jinming Nian \\
Santa Clara University \\
\texttt{jnian@scu.edu} \\\And
Zhiyuan Peng \\
Santa Clara University \\
\texttt{zpeng@scu.edu} \\\And
Hongwei Shang \\
Walmart Global Tech \\
\texttt{hongwei.shang@walmart.com} \\\AND
Dae Hoon Park \\
Walmart Global Tech \\
\texttt{dae.hoon.park@walmart.com} \\\And
Yi Fang \\
Santa Clara University \\
\texttt{yfang@scu.edu}
}

\begin{document}
\maketitle
\begin{abstract}

Automatic prompt optimization reduces manual prompt engineering, but relies on task performance measured on a small, often randomly sampled evaluation subset as its main source of feedback signal. Despite this, how to select that evaluation subset is usually treated as an implementation detail. We study evaluation subset selection for prompt optimization from a principled perspective and propose \ours\footnote{\hyperlink{https://github.com/jmnian/SESS}{https://github.com/jmnian/SESS}}, a submodular evaluation subset selection method. We frame selection as maximizing an objective set function and show that, under mild conditions, it is monotone and submodular, enabling greedy selection with theoretical guarantees. Across GSM8K, MATH, and GPQA-Diamond, submodularly selected evaluation subsets can yield better optimized prompts than random or heuristic baselines. 

\end{abstract}

\section{Introduction}
\label{sec:introduction}
Large language models (LLMs) are highly sensitive to how a task is described in natural language, which makes instruction prompts crucial for real deployments~\cite{10.1145/3411763.3451760}. Small changes to wording or formatting can lead to large swings in accuracy and behavior~\cite{sclar2024quantifying,errica-etal-2025-wrong}, while the right prompting pattern can unlock capabilities such as multi-step reasoning~\cite{10.5555/3600270.3602070,10.5555/3600270.3601883}. In practice, however, prompt engineering is often a slow trial-and-error process. 

\begin{figure}[t]
  \includegraphics[width=\columnwidth]{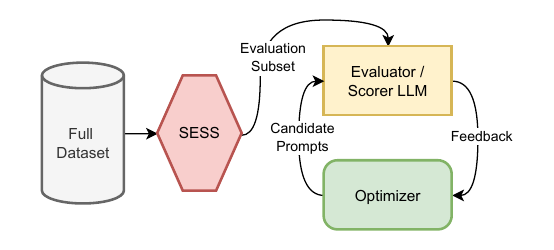}
  \caption{The general framework of APO. SESS replaces random or heuristic subset selection with a principled submodular evaluation subset selection module.}
  \label{fig:prompt_optimization_structure}
\end{figure}

To reduce manual effort, a growing line of work studies \textbf{automatic prompt optimization (APO)}. As illustrated in Figure~\ref{fig:prompt_optimization_structure}, a typical pipeline alternates between an LLM-based optimizer that proposes one or more candidate prompts, and an evaluator, which provide feedback in the form of evaluation scores or textual critiques. This feedback is then used to guide the next round of candidate prompt generation. For example, OPRO~\cite{opro} stores each evaluated prompt together with its score and uses the top-scoring prompts as in-context examples for subsequent optimization rounds. ProTeGi~\cite{pryzant2023automatic} and TextGrad~\cite{yuksekgonul2025optimizing} use gradient-inspired methods that treat LLM-generated natural language critiques as a proxy feedback for prompt optimization. More broadly, evolutionary search methods have also been used to iteratively improve algorithms using both evaluation score and critiques as feedback~\cite{novikov2025alphaevolvecodingagentscientific, lange2025shinkaevolveopenendedsampleefficientprogram}. Across methods, the shared pattern is: the optimizer proposes candidates, and evaluation feedback determines which candidates are retained for further optimization. 

Evaluating every candidate prompt on the full evaluation dataset is often infeasible; most APO methods score prompts on a randomly sampled evaluation subset~\cite {opro,pryzant2023automatic}. Because the optimizer only sees scores computed on the evaluation subset, this choice controls the noise of the feedback signal, how easily the optimizer overfits a few examples, and how stable prompt comparisons are across runs. This motivates a core yet underexplored \textbf{budgeted evaluation subset selection problem}: \textit{Given a fixed budget, how can we select an evaluation subset that best supports prompt optimization and leads to high-performing prompts on the target distribution?} Recent work has started to explore this direction. For instance, IPOMP~\cite{dong-etal-2025-model} selects evaluation data using a two-stage procedure based on semantic clustering and boundary analysis, followed by refinement using real-time model performance. Still, principled objectives with clear guarantees for evaluation subset selection specifically for prompt optimization remain limited.

To address this issue, we propose \ours, a submodular evaluation subset selection method for prompt optimization. We evaluate \ours on GSM8K~\cite{cobbe2021trainingverifierssolvemath}, MATH~\cite{hendrycksmath2021}, and GPQA-Diamond~\cite{rein2024gpqa}, where prompts optimized using submodularly selected evaluation subsets can outperform random and heuristic baselines. Our key contributions include: (1) we introduce \ours, a principled method to selecting evaluation subsets for APO, formulated as a budgeted set maximization problem; (2) we show that the proposed objectives are monotone and submodular under mild conditions, which allows greedy selection with approximation guarantees; (3) we present a unified framework that generalizes multiple subset selection strategies through different objective definitions, making optimization feedback an explicit modeling choice rather than an implementation detail; and (4) we show gains on GSM8K, MATH, and GPQA-Diamond, where \ours yields better optimized prompts than random or heuristic baselines.

\section{Related Work}
\paragraph{Subset Selection.}
Recent work on benchmark compression and coreset selection~\cite{DBLP:conf/acl/YuanZF0WSTPH025, vivek-etal-2024-anchor, DBLP:conf/iclr/KipnisVBS25, wang2025rethinking} focus on preserving full-benchmark scores or model ranking using a small subset. In contrast, we study evaluation subset selection for APO, where the goal is to maximize the target distribution performance of the optimized prompt. A subset that closely matches the full evaluation set is therefore not necessarily optimal for this goal. IPOMP~\cite{dong-etal-2025-model} addresses this problem using a heuristic two-stage procedure based on semantic clustering, boundary analysis, and model performance. While effective, IPOMP does not provide an explicit optimization objective or theoretical guarantees. Our approach complements this line of work by formulating evaluation subset selection as a budgeted set maximization problem with monotone submodular objectives, enabling greedy selection with approximation guarantees.

\section{Methodology}
\label{sec:method}

\subsection{Evaluation subset selection} 
Let $D=\{x_1,\dots,x_N\}$ be a pool of candidate evaluation examples. Automatic prompt optimization such as OPRO iteratively proposes prompts and keeps those that score well on an evaluation set. Since scoring each prompt on the full pool $D$ is often infeasible, we select a budgeted subset $S\subseteq D$ with $|S|\le k$ to serve as the optimization feedback.

We frame evaluation subset selection as the following budgeted set maximization problem:
\begin{equation}
S^* \in \arg\max_{S\subseteq D,\; |S|\le k} \mathcal{F}(S),
\label{eq:subset_opt}
\end{equation}
where $\mathcal{F}:2^D\to \mathbb{R}_{\ge 0}$ measures how suitable $S$ is for guiding prompt optimization. Exact maximization is NP-hard for many natural choices of $\mathcal{F}$, but when $\mathcal{F}$ is non-negative, monotone, and submodular~\cite{fujishige2005submodular}, the greedy algorithm achieves a $(1-1/e)$ approximation under the cardinality constraint~\cite{DBLP:journals/mp/NemhauserWF78}. Our goal is therefore to design useful evaluation objectives $\mathcal{F}$ that match the needs of prompt optimization and belong to this monotone submodular family. We solve Eq.~\ref{eq:subset_opt} once prior to optimization (static selection). We hypothesize that optimization feedback is primarily determined by intrinsic instance properties (e.g., representativeness or difficulty), allowing a fixed subset to provide a stable and efficient signal without the overhead of dynamic re-selection.

\subsection{Submodular Evaluation Objectives}
We consider two goals for evaluation subsets:
(1) \textbf{representativeness}: cover the diversity of the pool so the feedback reflects the full task distribution;
(2) \textbf{difficulty awareness}: emphasize examples that the scorer model finds hard, since these are often the most informative for comparing prompts. We propose three subset selection objectives: representative, least confident, and confidence-weighted representative. 

In the remainder of the paper, we refer to the greedy solution of each objective as a specific instance of \ours: \textbf{SESS-rep} for the representative objective $\mathcal{F}_{\mathrm{rep}}$, \textbf{SESS-lc} for least-confidence selection using likelihood-based confidence, \textbf{SESS-vlc} for least-confidence selection using verbalized confidence, both described by $\mathcal{F}_{\mathrm{lc}}$, and \textbf{SESS-wrep} for the confidence-weighted representative objective $\mathcal{F}_{\mathrm{wrep}}$. Each method selects a subset $S$ of size $k$ using the greedy procedure described in Section~\ref{sec:greedy}. 

\paragraph{Representative subset.}
We cast each example into a vector representation and define $\mathrm{sim}(i,j)$ by cosine similarity. To satisfy the non-negativity assumptions required by our analysis, we normalize cosine similarity as $\mathrm{sim}(i,j)=(1+\cos(i,j))/2\in[0,1]$. We then use a facility-location style objective to quantify how well a subset $S$ covers the full pool:
\begin{equation}
\mathcal{F}_{\mathrm{rep}}(S)\;:=\;\sum_{j\in D}\max_{i\in S}\mathrm{sim}(i,j).
\label{eq:facility_location}
\end{equation}
Intuitively, each example $j$ contributes its similarity to its nearest neighbor in the selected subset $S$, so larger $\mathcal{F}_{\mathrm{rep}}(S)$ indicates that $S$ contains good representatives for many regions of the pool. Under $\mathrm{sim}\ge 0$, $\mathcal{F}_{\mathrm{rep}}$ is monotone and submodular~\cite{DBLP:conf/nips/IyerB13}. Proofs are provided in Appendix~\ref{ap:proof}.

\paragraph{Least confident subset.}
Let $c(j)$ be a scalar confidence score for example $j$ computed by a fixed scorer model. We select the $k$ examples with smallest $c(j)$. This can be written as a modular objective
\begin{equation}
\mathcal{F}_{\mathrm{lc}}(S)\;:=\;\sum_{j\in S} (1-\tilde c(j)),
\label{eq:least_confident}
\end{equation}
where $\tilde c(j)\in[0,1]$ is a normalized confidence score of $c(j)$. We provide two variants: likelihood-based confidence and verbal confidence~\cite{tian-etal-2023-just}. These objectives capture difficulty but could cause redundancy among selected examples. Because the objective is modular and nonnegative, it is monotone and submodular.  

\paragraph{Confidence-weighted representative subset.}
To combine representativeness and difficulty awareness, we weight coverage to favor hard examples:
\begin{equation}
\mathcal{F}_{\mathrm{wrep}}(S)\;:=\;\sum_{j\in D} w(j)\cdot \max_{i\in S}\mathrm{sim}(i,j),
\label{eq:weighted_facility_location}
\end{equation}
where $w(j)\ge 0$ is an importance weight computed from the scorer model's likelihood-based confidence. In our implementation,
\begin{equation}
w(j) \;=\; (1-\lambda)\;+\;\lambda\cdot \bigl(1-\tilde c(j)\bigr),\  \lambda\in[0,1],
\label{eq:weight_def}
\end{equation}
so $\lambda=0$ gives us $\mathcal{F}_{\mathrm{rep}}(S)$ which recovers pure representativeness, while larger $\lambda$ increasingly concentrates coverage on low-confidence (hard) examples. Since $\mathcal{F}_{\mathrm{wrep}}$ is a nonnegative weighted sum of facility-location terms, it remains monotone submodular when $\mathrm{sim}\ge 0$ and $w(j)\ge 0$; we prove this in Appendix~\ref{ap:proof}. 

\subsection{Greedy selection}
\label{sec:greedy}
For any monotone submodular objective $\mathcal{F}$ above (in particular $\mathcal{F}_{\mathrm{rep}}$ and $\mathcal{F}_{\mathrm{wrep}}$), we apply the standard greedy algorithm starting from $S=\emptyset$:

\begin{equation}
x^* \leftarrow \arg\max_{x\in D\setminus S}\left[\mathcal{F}(S\cup\{x\})-\mathcal{F}(S)\right],
\label{eq:greedy}
\end{equation}
followed by the update $S \leftarrow S\cup\{x^*\}$, repeated until $|S|=k$. For monotone submodular $\mathcal{F}$ under the cardinality constraint, the resulting subset $S_{\mathrm{greedy}}$ enjoys the guarantee
\[
\mathcal{F}(S_{\mathrm{greedy}})\;\ge\;(1-1/e)\,\mathcal{F}(S^*)
\]
where $S^*$ is an optimal solution to Eq.~\eqref{eq:subset_opt}~\cite{DBLP:journals/mp/NemhauserWF78}. For modular objectives such as $\mathcal{F}_{\mathrm{lc}}$, $S_{\mathrm{greedy}}$ is obtained by sorting examples by uncertainty $(1-\tilde c(j))$ in descending order and selecting the top-$k$. 

\begin{table*}[t]
\centering
\small
\begin{tabular}{lccccccccc}
\toprule
\multirow{2}{*}{\textbf{Method}} &
\multicolumn{2}{c}{\textbf{GSM8K (EM)}} &
\multicolumn{2}{c}{\textbf{MATH (EM)}} &
\multicolumn{2}{c}{\textbf{GPQA-D (Acc)}} &
\multicolumn{3}{c}{\textbf{Avg}} \\
\cmidrule(lr){2-3}\cmidrule(lr){4-5}\cmidrule(lr){6-7}\cmidrule(lr){8-10}
& \textbf{Small} & \textbf{Large} & \textbf{Small} & \textbf{Large} & \textbf{Small} & \textbf{Large}
& \textbf{Small} & \textbf{Large} & \textbf{All} \\
\midrule
Base      & 82.9 & 82.9 & 73.2 & 73.2 & 29.9 & 29.9 & 62.0 & 62.0 & 62.0 \\
\midrule
Random    & 88.7 & 89.6 & 74.1 & 73.3 & \underline{34.3} & 35.9 & 65.7 & 66.3 & 66.0 \\
IPOMP     & 88.6 & \textbf{91.8} & 73.7 & 74.7 & 33.8 & 33.8 & 65.4 & 66.8 & 66.1 \\
Anchor-Points & 87.9 & 90.4 & 75.1 & \textbf{76.1} & 33.8 & \textbf{37.4} & 65.6 & \textbf{68.0} & 66.8 \\
\midrule
SESS-rep  & \textbf{91.9} & 89.9 & 73.2 & 70.3 & 32.3 & 29.8 & 65.8 & 63.3 & 64.6 \\
SESS-lc   & 90.8 & 90.8 & 73.9 & 75.3 & \underline{34.3} & 32.8 & 66.3 & 66.3 & 66.3 \\
SESS-vlc  & \underline{91.8} & \underline{91.0} & \underline{75.7} & \underline{75.7} & \underline{34.3} & \underline{36.9} & \underline{67.3} & \underline{67.9} & \textbf{67.6} \\
SESS-wrep & 90.0 & 90.8 & \textbf{76.1} & 73.6 & \textbf{37.4} & 34.8 & \textbf{67.8} & 66.4 & \underline{67.1} \\
\bottomrule
\end{tabular}
\caption{OPRO performance under different evaluation subset selection methods. We report Exact Match (EM) on GSM8K and MATH, and accuracy on GPQA-Diamond. \textbf{Small} and \textbf{Large} denote averages over small-budget (1\%, 1\%, 10\%) and large-budget (3.5\%, 3.5\%, 20\%) settings, respectively; \textbf{All} averages all six settings. \textbf{Bold} and \underline{underline} denote the best and second-best results in each column, respectively.}
\label{tab:main_results}
\end{table*}

\section{Experiment Setup} 
We evaluate how evaluation-subset selection in OPRO~\cite{opro} affect the final optimized prompt's test performance. OPRO iteratively proposes new instruction prompts using an optimizer LLM conditioned on previously tried prompts and their scores. Each candidate prompt is scored by running a scorer LLM on a fixed evaluation subset $S$, and the resulting score is the only feedback that guides optimization. We isolate the effect of subset selection by running the same OPRO procedure while varying how $S$ is chosen. After optimization, we select the highest-scoring candidate prompt on the evaluation subset and report its performance on the full test set.

\paragraph{Datasets.} We report Exact Match (EM) on GSM8K and MATH and accuracy on GPQA-Diamond. The evaluation budget is documented in Appendix~\ref{ap:sess-implementation-details}.

\paragraph{Subset selection methods.} We compare Base (``Let's solve the problem.'', no OPRO optimization), Random (OPRO default), IPOMP~\cite{dong-etal-2025-model}, Anchor-Points~\cite{vivek-etal-2024-anchor}, and our variants from Section~\ref{sec:method}: SESS-rep, SESS-lc (likelihood), SESS-vlc (verbal), and SESS-wrep. The implementation details of SESS can be found in Appendix~\ref{ap:sess-implementation-details}. Anchor-Points is designed to approximate full-benchmark evaluation with fewer examples, not specifically for prompt optimization. Although it is not proposed for prompt optimization, it serves as a strong baseline for selecting a coreset. 

\section{Results and Discussion}
Table~\ref{tab:main_results} reports the test performance on OPRO-optimized prompts when the evaluation subset is constructed by different selection methods. This directly tests our main claim that evaluation subset selection should not merely be an implementation detail, and that \ours provides a principled and effective way to shape the optimization signal and materially affect prompt optimization outcomes. 

\textbf{Overall, \ours variants are competitive with or outperform strong baselines, with the clearest gains under small evaluation budgets.} Among all methods, SESS-vlc achieves the best Avg(All) (67.6) and remains strong across datasets and budgets, while SESS-wrep achieves the best Avg(Small) (67.8). In contrast, Random and IPOMP perform notably worse under small budgets (65.7 and 65.4 Avg(Small)), indicating that careful subset construction matters most when evaluation resources are limited.

\textbf{Different objectives excel in different regimes, validating our unified objective family.} SESS-rep performs best on GSM8K at 1\% (91.9), suggesting that coverage-based objectives provide a strong signal on large, relatively homogeneous pools. SESS-wrep performs best on the most challenging small-budget setting, GPQA-Diamond 10\% (37.4). SESS-vlc is consistently strong across tasks and budgets, yielding the best Avg(All) and near-best Avg(Large).

\textbf{Notably, well-chosen small evaluation subsets can match or outperform larger budgets at a fraction of the compute.} For example, on GPQA-Diamond, SESS-wrep with a 10\% subset (20 questions) achieves 37.4\% accuracy, matching or exceeding several 20\% settings. Similar effects appear on GSM8K, where 1\% subsets already provide strong optimization signals. This suggests that increasing evaluation budget without improving subset quality yields diminishing returns: larger subsets often add redundant or low-signal examples, while carefully selected subsets concentrate feedback on representative yet hard cases, improving the signal-to-noise ratio seen by the optimizer.

Among baselines, Anchor-Points is the strongest competitor, achieving the best Avg(Large) (68.0) due to strong performance on MATH and GPQA-Diamond at large budgets. Random remains stable with Avg(All) of 66.0, while IPOMP performs well on GSM8K at 3.5\% (91.8) but does not generalize across datasets or budgets. Overall, \ours variants remain competitive across settings while offering explicit control over how optimization feedback is constructed.

\section{Conclusion} 
Across three benchmarks of varying difficulty and multiple evaluation budgets, \ours can yield stronger optimized prompts than random sampling and competitive heuristic baselines. This supports our core argument that evaluation subset selection is a core component of automatic prompt optimization, and framing it as monotone submodular set maximization provides both theoretical structure and practical gains. 

\clearpage
\newpage

\section{Limitations} 
Our experiments focus on a single prompt optimization pipeline OPRO with one optimizer LLM and one scorer LLM. While this setting allows controlled comparisons between subset selection methods, broader experiments across additional optimizer and scorer models would help clarify how sensitive the conclusions are to these settings. We also primarily report final test performance. A more detailed study of optimization dynamics, such as convergence speed and how quickly strong prompts are discovered, would provide a fuller picture of how subset selection shapes the optimization process. 

\section{Acknowledgment}
We used AI-based tools to assist with code drafting, debugging, and grammar checking during the research and writing process. All other aspects of the work were carried out by the authors.

\bibliography{custom}

\newpage
\appendix
\onecolumn

\section{Implementation Details}
\label{ap:sess-implementation-details}
For GSM8K and MATH, we consider two evaluation budgets: small budget with $|S|=75$ (1\% of the training split) and a large budget with $|S|=262$ (3.5\% of the training split). Since GPQA-Diamond has 198 questions, we treat the full set as the pool and select $|S|=20$ (10\%) under the small budget and $|S|=40$ (20\%) under the large budget
for prompt optimization. Final testing is performed on all 198 questions. All methods select $S$ once and keep it fixed throughout optimization, except IPOMP, as IPOMP is a two-stage method.

For SESS-rep, we build a question--question similarity matrix by mixing dense and lexical similarity. For dense similarity, we embed questions using \texttt{Qwen/Qwen3-Embedding-8B}~\cite{zhang2025qwen3embeddingadvancingtext} and compute a cosine-similarity matrix, which is then normalized to $[0,1]$. For lexical similarity, we compute a TF-IDF similarity matrix over the same questions, normalize it, and apply a square root transform to re-scale values. The final similarity is a convex combination
$
M=\alpha M_{\mathrm{dense}}+(1-\alpha)M_{\mathrm{tfidf}},
$
with $\alpha=0.7$ in all experiments. For SESS-lc, we score each example using the log-likelihood of the ground-truth answer conditioned on the question, normalized by answer length, and computed with a direct-answer prompt (Appendix~\ref{ap:lc_prompt}). For SESS-vlc, we use the verbalized-confidence prompt in Appendix~\ref{ap:vlc_prompt} and take the maximum reported probability as the confidence score. For SESS-wrep, we set the difficulty weight to $\lambda=0.5$ in Eq.~\eqref{eq:weight_def}.

We run OPRO for 100 steps with 7 candidates per step. We use \texttt{gpt-oss-120B} (4-bit)~\cite{openai2025gptoss120bgptoss20bmodel} as the optimizer LLM (temperature 1.0) and \texttt{Qwen/Qwen2.5-7B-Instruct}~\cite{qwen2025qwen25technicalreport} as the scorer LLM (temperature 0.0). Inference uses vLLM~\cite{10.1145/3600006.3613165}. We repeat each experiment 2 times and report the average. On 8 $\times$ NVIDIA A100-80GB GPUs, one full OPRO run takes approximately one hour. 



\section{Prompts}

\subsection{Verbal Confidence Prompt}
\label{ap:vlc_prompt}
\begin{figure}[H]
    \centering
    \begin{mdframed}[
        linecolor=black!60,
        linewidth=1pt,
        roundcorner=10pt,
        backgroundcolor=gray!5,
        shadow=true,
        shadowsize=5pt,
        shadowcolor=black!40,
        skipabove=10pt,
        skipbelow=10pt,
        innertopmargin=10pt,
        innerbottommargin=10pt,
        innerleftmargin=10pt,
        innerrightmargin=10pt
    ]
    Provide your 4 best guesses and the probability that each is correct (0.0 to 1.0) for the following question. Give ONLY the guesses and probabilities, no other words or explanation. For example: \\[0.3em]

    G1: <first most likely guess, as short as possible; not a complete sentence, just the guess!> \\
    P1: <the probability between 0.0 and 1.0 that G1 is correct, without any extra commentary whatsoever; just the probability!> \\

    G2: <second most likely guess, as short as possible; not a complete sentence, just the guess!> \\
    P2: <the probability between 0.0 and 1.0 that G2 is correct, without any extra commentary whatsoever; just the probability!> \\

    G3: <third most likely guess, as short as possible; not a complete sentence, just the guess!> \\
    P3: <the probability between 0.0 and 1.0 that G3 is correct, without any extra commentary whatsoever; just the probability!> \\

    G4: <fourth most likely guess, as short as possible; not a complete sentence, just the guess!> \\
    P4: <the probability between 0.0 and 1.0 that G4 is correct, without any extra commentary whatsoever; just the probability!> \\[0.5em]

    The question is: \\
    \{question\}
    \end{mdframed}
    \caption{Verbal confidence elicitation prompt}
    \label{fig:verbal_conf_prompt}
\end{figure}

\subsection{Log-likelihood Confidence Prompt}
\label{ap:lc_prompt}
\begin{figure}[H]
    \centering
    \begin{mdframed}[
        linecolor=black!60,
        linewidth=1pt,
        roundcorner=10pt,
        backgroundcolor=gray!5,
        shadow=true,
        shadowsize=5pt,
        shadowcolor=black!40,
        skipabove=10pt,
        skipbelow=10pt,
        innertopmargin=10pt,
        innerbottommargin=10pt,
        innerleftmargin=10pt,
        innerrightmargin=10pt
    ]
    Directly give the choice A or B or C or D: \{question\} \\[0.3em]
    Answer: \{answer\}
    \end{mdframed}
    \caption{Multiple-choice answer prompt used for likelihood-based confidence.}
    \label{fig:mc_likelihood_prompt}
\end{figure}
\begin{figure}[H]
    \centering
    \begin{mdframed}[
        linecolor=black!60,
        linewidth=1pt,
        roundcorner=10pt,
        backgroundcolor=gray!5,
        shadow=true,
        shadowsize=5pt,
        shadowcolor=black!40,
        skipabove=10pt,
        skipbelow=10pt,
        innertopmargin=10pt,
        innerbottommargin=10pt,
        innerleftmargin=10pt,
        innerrightmargin=10pt
    ]
    Directly give the numeric answer to the following question: \{question\} \\[0.3em]
    Answer: \{answer\}
    \end{mdframed}
    \caption{Numeric/free-form answer prompt used for likelihood-based confidence.}
    \label{fig:numeric_likelihood_prompt}
\end{figure}

\section{Proofs}
\label{ap:proof}

\begin{proof}[Proof for submodularity of $\mathcal{F}_{\mathrm{rep}}$ and $\mathcal{F}_{\mathrm{wrep}}$]
A set function $\mathcal{F}$ is submodular if it satisfies the diminishing-returns property: for all
$A \subseteq B \subseteq D$ and all $x \in D \setminus B$,
\begin{equation}
\mathcal{F}(A \cup \{x\}) - \mathcal{F}(A) \ge \mathcal{F}(B \cup \{x\}) - \mathcal{F}(B).
\label{eq:diminishing_returns}
\end{equation}

\textbf{Submodularity of $\mathcal{F}_{\mathrm{rep}}$.}
Fix any $A \subseteq B \subseteq D$ and any $x \in D \setminus B$. For each $j \in D$, define
\[
a_j := \max_{i \in A}\mathrm{sim}(i,j), \qquad
b_j := \max_{i \in B}\mathrm{sim}(i,j).
\]
Since $A \subseteq B$, we have $a_j \le b_j$ for all $j \in D$. The marginal gain contributed by index $j$
when adding $x$ to $A$ is
\begin{align}
\delta_j(x \mid A)
&:= \max_{i \in A \cup \{x\}} \mathrm{sim}(i,j) - \max_{i \in A}\mathrm{sim}(i,j) \notag \\
&= \max\bigl(\mathrm{sim}(x,j), a_j\bigr) - a_j \notag \\
&= \max\bigl(\mathrm{sim}(x,j) - a_j,\, 0\bigr).
\label{eq:margin_gain}
\end{align}
Similarly,
\[
\delta_j(x \mid B) = \max\bigl(\mathrm{sim}(x,j) - b_j,\, 0\bigr).
\]
Because $a_j \le b_j$, we have $\mathrm{sim}(x,j)-a_j \ge \mathrm{sim}(x,j)-b_j$, and since the map
$t \mapsto \max(t,0)$ is non-decreasing,
\[
\delta_j(x \mid A) \ge \delta_j(x \mid B)
\quad\text{for all } j \in D.
\]
Summing over $j \in D$ yields
\[
\mathcal{F}_{\mathrm{rep}}(A\cup\{x\})-\mathcal{F}_{\mathrm{rep}}(A)
= \sum_{j\in D}\delta_j(x\mid A)
\ge \sum_{j\in D}\delta_j(x\mid B)
= \mathcal{F}_{\mathrm{rep}}(B\cup\{x\})-\mathcal{F}_{\mathrm{rep}}(B),
\]
which is exactly Eq.~\eqref{eq:diminishing_returns}. Therefore, $\mathcal{F}_{\mathrm{rep}}$ is submodular.

\textbf{Submodularity of $\mathcal{F}_{\mathrm{wrep}}$.}
Assume $w(j)\ge 0$ for all $j \in D$. Define the weighted marginal gains
\[
\Delta_j(x \mid A) := w(j)\cdot \delta_j(x \mid A),
\qquad
\Delta_j(x \mid B) := w(j)\cdot \delta_j(x \mid B).
\]
Since $w(j)\ge 0$ and $\delta_j(x \mid A)\ge \delta_j(x \mid B)$, we have
$\Delta_j(x \mid A)\ge \Delta_j(x \mid B)$ for all $j \in D$. Summing over $j \in D$ gives
\begin{align*}
\mathcal{F}_{\mathrm{wrep}}(A\cup\{x\})-\mathcal{F}_{\mathrm{wrep}}(A)
&= \sum_{j\in D} w(j)\,\delta_j(x\mid A)
\;\ge\; \sum_{j\in D} w(j)\,\delta_j(x\mid B) \\
&= \mathcal{F}_{\mathrm{wrep}}(B\cup\{x\})-\mathcal{F}_{\mathrm{wrep}}(B),
\end{align*}
so $\mathcal{F}_{\mathrm{wrep}}$ is submodular.
\end{proof}

\begin{proof}[Proof for monotonicity of $\mathcal{F}_{\mathrm{rep}}$ and $\mathcal{F}_{\mathrm{wrep}}$]
We prove monotonicity in the standard sense: for any $S \subseteq T \subseteq D$,
$\mathcal{F}(S) \le \mathcal{F}(T)$.

\textbf{Monotonicity of $\mathcal{F}_{\mathrm{rep}}$.}
Fix any $S \subseteq T \subseteq D$. For any $j \in D$, since $S \subseteq T$,
\[
\max_{i \in S}\mathrm{sim}(i,j) \;\le\; \max_{i \in T}\mathrm{sim}(i,j).
\]
Summing over $j \in D$ yields
\[
\mathcal{F}_{\mathrm{rep}}(S)
= \sum_{j \in D}\max_{i \in S}\mathrm{sim}(i,j)
\;\le\;
\sum_{j \in D}\max_{i \in T}\mathrm{sim}(i,j)
= \mathcal{F}_{\mathrm{rep}}(T),
\]
so $\mathcal{F}_{\mathrm{rep}}$ is monotone.

\textbf{Monotonicity of $\mathcal{F}_{\mathrm{wrep}}$.}
Assume $w(j)\ge 0$ for all $j \in D$. Using the same pointwise inequality as above and multiplying by $w(j)\ge 0$ preserves the inequality:
\[
w(j)\cdot \max_{i \in S}\mathrm{sim}(i,j) \;\le\; w(j)\cdot \max_{i \in T}\mathrm{sim}(i,j).
\]
Summing over $j \in D$ gives
\[
\mathcal{F}_{\mathrm{wrep}}(S)
= \sum_{j \in D} w(j)\cdot \max_{i \in S}\mathrm{sim}(i,j)
\;\le\;
\sum_{j \in D} w(j)\cdot \max_{i \in T}\mathrm{sim}(i,j)
= \mathcal{F}_{\mathrm{wrep}}(T),
\]
hence $\mathcal{F}_{\mathrm{wrep}}$ is monotone.
\end{proof}

\end{document}